\pdfoutput=1

\documentclass[11pt]{article}

\usepackage{acl}

\usepackage{times}
\usepackage{latexsym}

\usepackage[T1]{fontenc}

\usepackage[utf8]{inputenc}

\usepackage{microtype}

\usepackage{amsmath,amsfonts,amssymb}
\usepackage{pifont}
\usepackage{graphicx}
\usepackage{booktabs}
\usepackage{multirow}
\usepackage[referable]{threeparttablex}
\usepackage{xspace}
\usepackage{xcolor}
\usepackage{tablefootnote}
\usepackage[title]{appendix}

\usepackage{bm}
\usepackage{siunitx}

\newcommand{\cmark}{\ding{51}}%
\newcommand{\xmark}{\ding{55}}%

\newcommand{\ours}{VideoOFA}

\def\vx{{\mathbf{x}}}
\def\vy{{\mathbf{y}}}

\title{\ours: Two-Stage Pre-Training for Video-to-Text Generation}

\author{
Xilun Chen \And Lili Yu \And Wenhan Xiong \AND Barlas Oğuz \And Yashar Mehdad \And Wen-tau Yih \AND
\textnormal{Meta AI}\\
\texttt{\{xilun,liliyu,xwhan,barlaso,mehdad,scottyih\}@meta.com}
}

\begin{document}
\maketitle
\begin{abstract}
We propose a new \emph{two-stage pre-training} framework for video-to-text generation tasks such as video captioning and video question answering:
A generative encoder-decoder model is first jointly pre-trained on massive \emph{image-text} data to learn fundamental vision-language concepts, and then adapted to video data in an \emph{intermediate video-text pre-training} stage to learn video-specific skills such as spatio-temporal reasoning.
As a result, our \ours{} model achieves new state-of-the-art performance on four Video Captioning benchmarks, beating prior art by an average of 9.7 points in CIDEr score.
It also outperforms existing models on two open-ended Video Question Answering datasets, showcasing its generalization capability as a universal video-to-text model.
\end{abstract}

\section{Introduction}\label{sec:intro}

\begin{figure*}
    \centering
    \includegraphics[width=\linewidth]{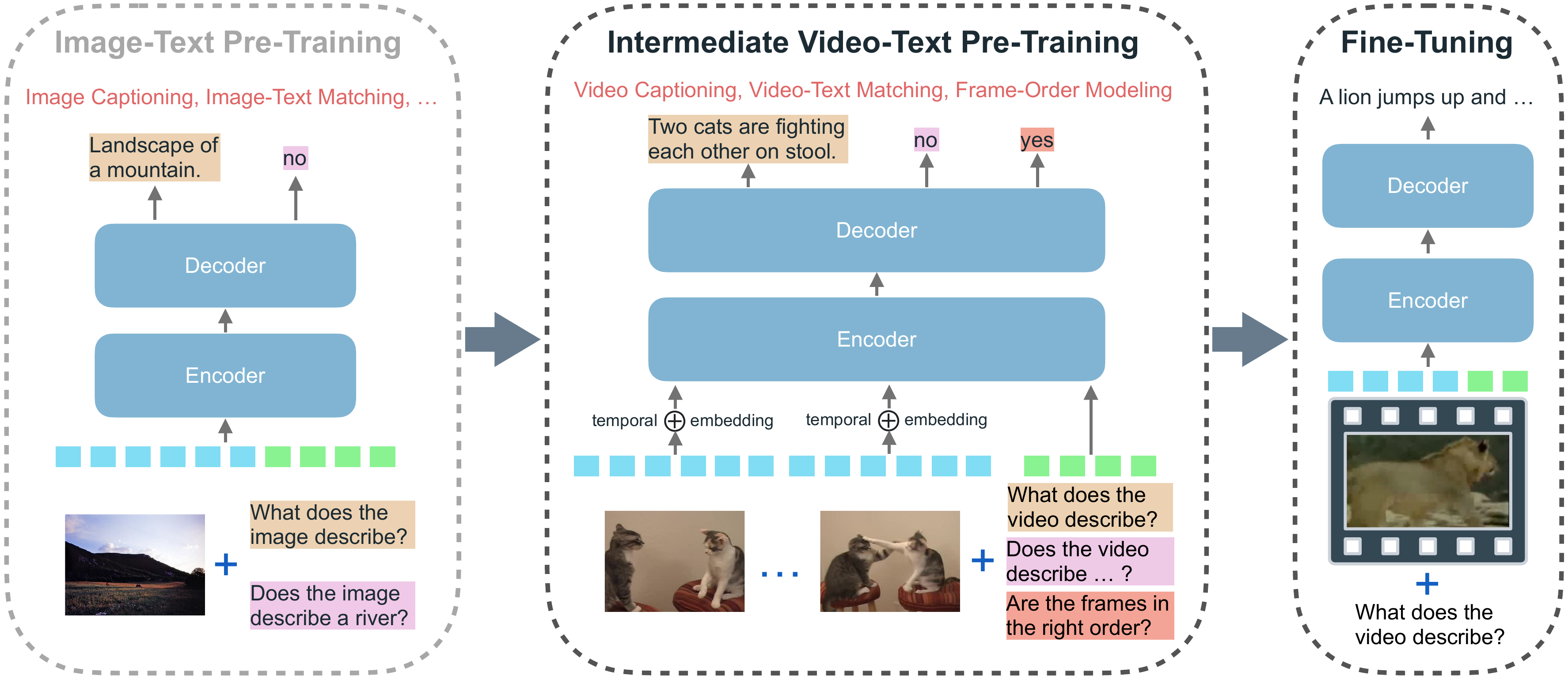}
    \caption{Overview of our two-stage pre-training framework for video-to-text generation. The generative encoder-decoder model is first pre-trained on the abundant image-text data to learn fundamental visual-language representations, and then further pre-trained on video-text data to learn video-specific concepts such as temporal reasoning. In practice, we adopt state-of-the-art generative image backbones such as OFA and focus on the latter stage.}
    \label{fig:framework}
\end{figure*}

Video-to-text generation is a family of video-language tasks that requires the model to produce a suitable text response given the multimodal video input.
A yardstick of success on video-to-text generation is the video captioning task~\citep{kojima-etal-2002-natural,li-etal-iccv2015-describing} that aims to generate a text description summarizing a given video.
Another example is video question answering~\citep[VideoQA,][]{zhu2017uncovering,zeng2017leveraging}, where the model aims to produce a textual answer given a video and a question.
Compared to other video-language tasks such as video classification or retrieval, video-to-text generation creates unique challenges in that the model needs not only to learn accurate and fine-grained video representations, but to generate coherent natural language results.

In recent years, considerable progress has been made in video-to-text generation, thanks in part to the Transformer model~\cite{vaswani2017attention} and self-supervised pre-training~\citep{devlin-etal-2019-bert}.
Earlier transformer-based models~\citep{Luo2020UniVL,li-etal-2020-hero} typically rely on \emph{video-language pre-training} that pre-trains a randomly initialized encoder-decoder transformer on large-scale video-text data.
More recent models found that better performance can be achieved if the (whole or part of) model is initialized from a well-trained image understanding backbone model.
Such backbone models are often trained on web-scale image-text data, resulting in higher-quality vision-language representations.
When these superior representations are adapted to video tasks, higher performance can be achieved compared to only training on video data.
For example, CLIP4Caption~\cite{clip4caption} achieved state-of-the-art performance by adapting the representations from a highly successful image-text model, CLIP~\citep{pmlr-v139-radford21a}, to video captioning.

\begin{table*}[ht]
\footnotesize
\centering
\begin{threeparttable}
\begin{tabular} {l rrrrr}
\toprule
Model    & Encoder Init & Decoder Init & Video Frames Handling & Image-Text PT & Video-Text PT \\ 
\midrule
UniVL\tnote{1}   & \color{lightgray} random & \color{lightgray} random (3 layers)  & \color{lightgray} frame-level pooling  & \color{lightgray} \xmark & \cmark \\
HERO\tnote{2}    & \color{lightgray} random & \color{lightgray} random (2 layers) & \color{lightgray} frame-level pooling & \color{lightgray} \xmark & \cmark \\
CLIP4Caption\tnote{3}    & CLIP  & \color{lightgray} UniVL (3 layers)  & \color{lightgray} frame-level pooling & \color{lightgray} encoder-only & \color{lightgray} \xmark \\
SwinBERT\tnote{4}    & VidSwin   & \color{lightgray} random    & 3D local attention & \color{lightgray} \xmark & \cmark \\
MV-GPT\tnote{5}      & ViViT & GPT2 \textcolor{lightgray}{(disjoint)}  & 3D local attention    & \color{lightgray} \xmark & \cmark \\
LAVENDER\tnote{6}    & VidSwin   & BERT \textcolor{lightgray}{(disjoint)} & 3D local attention & \color{lightgray} \cmark\kern-1.1ex\raisebox{.7ex}{\rotatebox[origin=c]{125}{--}} & \cmark \\
HiTeA\tnote{7}       & MViT  & BERT \textcolor{lightgray}{(disjoint)} & full token-level attention    & \color{lightgray} \xmark & \cmark \\
GIT\tnote{8}     & GIT   & GIT   & fusion-in-decoder & \cmark & \color{lightgray} \xmark \\
\midrule
\ours{} (ours) & OFA   & OFA   & full token-level attention & \cmark & \cmark \\
\bottomrule
\end{tabular}

\begin{tablenotes}[para,raggedright]
\item[1] \citep{Luo2020UniVL}
\item[2] \citep{li-etal-2020-hero}
\item[3] \citep{clip4caption}
\item[4] \citep{Lin_2022_CVPR_SwinBERT}
\item[5] \citep{mv-gpt}
\item[6] \citep{lavender}
\item[7] \citep{hitea}
\item[8] \citep{wang2022git}
\end{tablenotes}

\end{threeparttable}

\caption{An overview of existing approaches on video-to-text generation. Fields in \textcolor{lightgray}{gray} indicate sub-optimal choices that may limit the model capacity. ``Disjoint'' means the encoder and decoder are pre-trained separately.}
\label{tab:methods-survey}
\end{table*}

Most of these image \emph{backbone} models, however, are encoder-only models that focus on learning better image representations.
Many recent video-to-text models, therefore, combine the powerful image encoders with \emph{shallow}, \emph{random}, or \emph{separately trained} text decoders, limiting the capacity of the final model.
As summarized in Table~\ref{tab:methods-survey}, several existing models combine visual encoders such as VidSwin~\cite{vidswin} with independently pre-trained text decoders such as BERT~\cite{devlin-etal-2019-bert} or GPT2~\cite{radford2019language}.
One exception is GIT~\cite{wang2022git}, a very recent work that pre-trains the entire encoder-decoder model on large-scale image-text data.
It, however, simply fine-tunes the generative image-text model on downstream evaluation datasets, which is not an ideal way to adapt the image backbone to video understanding due to the limited data available during fine-tuning.
In particular, video understanding is more challenging than image understanding as it involves an additional temporal dimension.
Additional pre-training on video data may be needed to properly adapt an image backbone to video understanding by learning to model temporal reasoning, identify important moments of the video, etc.

Therefore, loosely inspired by task-adaptive pre-training~\cite{gururangan-etal-2020-dont,aghajanyan-etal-2021-muppet}, we in this work propose a new \textbf{two-stage pre-training} framework for learning video-to-text generation models (Figure~\ref{fig:framework}):
First, a generative model is pre-trained on massive image-text data (\emph{Image-Text Pre-Training}) to learn fundamental vision-language representations.\footnote{In practice, state-of-the-art generative image-text models such as OFA~\cite{wang-etal-2022-ofa} can be adopted instead of training from scratch.}
Then the architecture needs to be adapted to handle video data (e.g.\ adding temporal embeddings), ideally without introducing shallow components or too many non-pretrained parameters.
Next, we propose an intermediate \emph{Video-Text Pre-Training} stage where we further pre-train the generative image backbone on large-scale video-text data to learn video-specific skills such as temporal reasoning, before the model is fine-tuned on downstream evaluation datasets.

In summary, our contributions are as follows:
\vspace{-0.2cm}
\begin{itemize}
\itemsep-0.4em 
\item We propose a new two-stage training paradigm for generative video-language tasks.
\item We study various \emph{architectures} and \emph{pre-training strategies} under our framework and provide new insights.
\item The resulting \ours{} model achieves new state-of-the-art performance on four video captioning benchmarks, leading by an average of \textbf{9.7 points} in CIDEr score.
\item \ours{} also beats existing models on two video question answering datasets, despite not pre-trained on VideoQA data, demonstrating its generalizability.
\end{itemize}
\section{Method}\label{sec:method}
In this section, we detail our two-stage pre-training framework and propose our \ours{} model for video-to-text generation.

\subsection{The Image-Text Backbone}\label{sec:method:backbone}
As argued in Section~\ref{sec:intro}, a video-to-text model should first be pre-trained on the more abundant image-text data to learn fundamental visual-language representations.
Fortunately, there have been a multitude of recent works on large-scale image-text pre-training for generative vision-language tasks~\citep[][\emph{inter alia}]{vl-t5,unitab}.
In this work, we adopt OFA-large~\cite{wang-etal-2022-ofa} as the backbone of our model, which achieves state-of-the-art performance on generative image-text tasks such as image captioning and visual question answering.
We thus focus on adapting an image backbone to video-to-text generation, and study a backbone-agnostic two-stage pre-training framework that is compatible with more effective generative image backbone models that may come out in the future.

We briefly introduce the OFA model in this section and refer the readers to the original paper~\cite{wang-etal-2022-ofa} for more details.
OFA is a multimodal generative model that unifies the handling of image and text data in a single seq2seq transformer.
An input image $\vx^i\in\mathbb{R}^{H\times W\times C}$ is split into $P$ fixed-size patches, each of which is projected into an $H$-dimensional embedding vector.\footnote{For simplicity, we use \emph{patches} and \emph{tokens} interchangeably: A patch is also an image token.}
Similarly, a text input $\vx^t = w_1, \dots, w_n$ is first tokenized into BPE tokens and then embedded to $H$-dimensional vectors.
The image and text tokens can thus be concatenated and fed together to the encoder.
On the decoder side, a single unified output vocabulary is used for all modalities, where images are discretized into tokens using sparse coding~\cite{vqgan}.

OFA is trained on massive image-text data and multiple tasks, making it a flexible framework for a variety of vision-language tasks.
For example, \emph{Image Captioning} takes the input image as well as a text instruction ``What does the image describe?'' and generates a text description of the image;
\emph{Visual Question Answering} reads in the image and the text question and predicts the answer;
\emph{Conditional Image Generation} takes a text description such as ``a chair with three legs'' and produces an output image.
In this work, we focus on text output on the decoder side to adapt the OFA backbone to multimodal video-to-text generation, and leave the exploration of more general multimodal generative video models (e.g.\ video-to-image and video-to-video) for future work.

\subsection{Image-to-Video Adaptation}\label{sec:method:arch}
In this section we discuss the architecture changes needed to adapt the image-text backbone to video-text tasks.
Some early works rely on \emph{frame-level pooling} that represents each video frame into a single hidden vector in the encoder~\cite{Luo2020UniVL,clip4caption}.
However, compressing the entire video frame into a single vector inevitably limits the representational power of the model and many useful details may be lost.
Recent models, therefore, tend to adopt finer-grained tokens for more accurate representations of the video.
On the other hand, as the memory use of a transformer model scales quadratically with the number of tokens, several methods have been studied to reduce the memory footprint, mainly by altering the attention mechanism.

\paragraph{Video Tokenization}
One popular approach is to view a video as a 3D tensor of pixels $\vx^v\in \mathbb{R}^{T\times H\times W\times C}$, where $T$ is the additional temporal dimension.
Similarly to how an image is divided to non-overlapping 2D patches, some previous works divide a video into a number of non-overlapping 3D ``patches'' (or tubelets), each of which is treated as an input token~\cite{vivit,vidswin}.
This method, however, defines tokens differently as in the image backbone, which we argue is not ideal for adapting the well pre-trained backbone model to video-text tasks.
We instead propose to preserve the image tokenization in the backbone model, and simply formulate the video as a sequence of images (frames). 

\paragraph{Attention Mechanism}
As the total number of tokens in a video is much greater than an image, various methods have been explored to reduce the memory use of the transformer model.
Other than the frame-level pooling that reduces each frame into a single vector used in early work, another popular technique is factorized or local attention~\cite{vivit,vidswin}.
These methods exploit the spatio-temporal locality of the video data and assumes that pixels that are closer to each other spatially and/or temporally are more likely to be correlated.
They thus limit the attention module in a particular transformer layer to only attend to a subset of tokens based on some spatio-temporal distance metrics.
For instance, factorized attention performs attention along the spatial dimension first followed by the temporal dimension, while 3D local attention only attends to a 3D spatio-temporal window around each token.

In this work, we propose two straightforward attention mechanisms without having to design specialized heuristics, one optimized for accuracy while the other for efficiency.\footnote{Similar approaches have also been independently explored in contemporaneous works~\cite{wang2022git,hitea}}
Our goal is to introduce as little change as possible to the image backbone to retain its vision-language representations learned through image-text pre-training, while adding temporal reasoning capability required for video-text tasks.

In our full \ours{} model, we simply sparsely sample frames from an input video and concatenate the image tokens from each video frame to form a single sequence of tokens. 
We then add an additional positional embedding to indicate which frame each token is from, and these temporal embeddings are zero-initialized learnable parameters added on top of the spatial position embeddings.
The attention mechanism is not modified and it allows each token to freely attend to all other tokens for maximum interaction and expressiveness.

We also propose a more memory-efficient variant of the model, \ours{}-FiD, inspired by the fusion-in-decoder idea~\cite{izacard-grave-2021-leveraging}.
In this variant, the encoder independently encode each video frame, and the encoder output from each frame is then concatenated before fed into the decoder.
Temporal embeddings are optional: When enabled, they are added to the encoder output and used by the decoder.
Intuitively, the image encoder learns hidden representations for each frame, and the decoder fuses these representations together to jointly reason over the entire video and generate the text output.
As a comparison, the full \ours{} model can fuse information from different frames in the entire model, where the encoder learns spatio-temporal representations instead of spatial-only.
\ours{}-FiD is equivalent to applying spatial-only local attention to the encoder while allowing full self- and cross-attention in the decoder.
Compared to other local attention mechanisms, our \ours{}-FiD approach keeps the image encoder intact and can fully leverage the image representations from the backbone model.

\subsection{Intermediate Video-Text Pre-Training}\label{sec:method:pft}
While the image-text backbone understands static visual concepts, such as objects and scenes, additional capabilities are required for a video-to-text model.
In particular, our \ours{} model needs to encode temporal information across multiple video frames to learn representations for actions, activities, events, etc., while the decoder also needs to be able to reason over multiple frames to generate appropriate text output.
For instance, the image contents of the video frames that correspond to ``picks up the box'' and ``drops off the box'' may be almost identical with only the order being different.
To acquire such new capabilities, we propose an intermediate video-text pre-training stage on top of the image-text pre-trained backbone model, before it is applied to downstream video-to-text tasks.

We investigate three video-text pre-training tasks for the intermediate pre-training (IPT) stage, namely \emph{video captioning}, \emph{video-text matching}, and \emph{frame order modeling} (see Figure~\ref{fig:framework}).
To unify all tasks under a single framework, we formulate all pre-training (and fine-tuning) tasks as sequence-to-sequence (seq2seq) problems~\cite{t5} following the OFA backbone.

\paragraph{Video Captioning}
Given a video-text pair $(\vx^v, \vy^t)$, where $\vy^t$ is a text description (caption) of the video, the input of the video captioning task consists of the video $\vx_v$ and a text instruction $\vx^t = \text{\emph{What does the video describe?}}$.
The model learns to predict $\vy^t$ given this input, and is trained using teacher forcing and the cross-entropy loss:
\begin{equation}
    \mathcal{L}_\theta (\vx^v,\vx^t, \vy^t) = -\log\mathbb{P}_\theta(\vy^t | \vx^v, \vx^t)
    \label{eqn:loss}
\end{equation}
where $\theta$ denotes the trainable model parameters.

\paragraph{Video-Text Matching}
For a given video-text pair $(\vx^v, \vy^t)$, the video-text matching task aims to differentiate $\vy^t$ from some random text $\hat{\vy}^t$ that does not describe the video.
The video-text matching task is typically modeled by a contrastive training objective~\cite{Luo2020UniVL,li-etal-2020-hero}, which is not congruent with our generative seq2seq framework.
We instead propose a seq2seq formulation where the input consists of the video $\vx^v$ and text prompt $\vx^t=\text{\emph{Does the video describe }} \vy^t \text{\emph{?}}$, and the model learns to generate a verbal ``yes'' or ``no'' depending on whether $\vx^v$ and $\vy^t$ are paired.
For each sample in the training data, we create a positive or negative sample with equal probability, by feeding the model with either $(\vx^v, \vy^t)$ or $(\vx^v, \hat{\vy}^t)$ where $\hat{\vy}^t$ is a caption randomly selected from the entire training corpus.

\paragraph{Frame Order Modeling}
We further propose the frame order modeling (FOM) task~\cite{li-etal-2020-hero} to explicitly teach the model temporal information.
We consider two variants of the FOM task, a \emph{generative} one and a \emph{contrastive} one.
In generative FOM, the input includes a corrupted video $\hat{\vx}^v$ where the order of the video frames are randomly shuffled\footnote{Only 25\% of the frames are shuffled in practice to control the difficulty of the task.}, as well as a text instruction $\vx^t=\text{\emph{What is the correct frame order in the video?}}$.
The model is then trained to predict a sequence of numbers indicating the correct order of frames $\vy^t=p_0, \dots, p_n$ where $p_i$ is the correct position (in the original video $\vx^v$) of the $i$-th frame in $\hat{\vx}^v$.

In contrastive FOM, for each video $\vx^v$ we create a positive or negative sample with equal probability. 
The positive sample has input $\vx^v$ and instruction $\vx^t=\text{\emph{Are the frames in the video in the correct order?}}$, and output $\vy^t=\text{\emph{yes}}$.
Similarly, the negative sample has the same instruction $\vx^t$, but a shuffled video $\hat{\vx}^v$, and the output $\vy^t=\text{\emph{no}}$.

\paragraph{Final Intermediate Pre-Training Strategy}
We conduct an empirical study on various IPT strategies to select the optimal solution (Section~\ref{sec:analysis:ipt_strategy}), and find that all three tasks are beneficial.
For FOM, contrastive FOM alone outperforms generative + contrastive FOM.
One possible reason is that training the model to directly generate the frame order, an artifact not seen in real tasks, can degrade the quality of the model.
On the other hand, contrastive FOM can learn to encode the temporal information in videos without training the model to generate undesirable artifacts.
As a result, we adpot video captioning, video-text matching, and contrastive FOM in our final model.
\section{Experiments}\label{sec:exp}
In this section, we first introduce the datasets used during our intermediate video-text pre-training.
We then present our main results on Video Captioning and Video Question Answering.

\subsection{Intermediate Pre-Training Datasets}\label{sec:exp:ipt_data}
A total of 2.2 million video-text pairs from the following three datasets are used for our intermediate pre-training.

\paragraph{Spoken Moments-in-Time~\citep[S-MiT,][]{Monfort_2021_CVPR_SMiT}}
S-MiT is a video description dataset with 500K video clips covering a broad range of events, each associated with a high-quality text description (caption).

\paragraph{Kinetics 700-2020~\citep[Kinetics,][]{kinetics700}}
Kinetics 700 contains 650K video clips of 700 types of human actions.
Each video is annotated with a class label indicating the type of human action, typically a word or short phrase such as \emph{zumba} or \emph{tasting wine}.
These class labels are used as the text descriptions for each video during IPT.

\paragraph{HowTo100M~\cite{miech19howto100m}}
HowTo100M is a collection of 1.2 million instruction videos for a variety of tasks and activities such as cooking, gardening, etc.
Each video is devided into smaller video clips, where each clip is provided with a narration available as subtitles that is either written manually or produced by an automatic speech recognition (ASR) system.
To facilitate training, we bring the size of the three datasets to the same order of magnitude by sampling one clip (and its associated subtitle) for each video in HowTo100M, resulting in a total of 1.2M video-text pairs.

In Section~\ref{sec:analysis:ipt_strategy}, we experiment with different choices of IPT data, and find that using all three datasets yield the best results.

\subsection{Video Captioning}\label{sec:exp:captioning}
\begin{table*}[ht]
\setlength{\tabcolsep}{0.25em}
\footnotesize
\centering
\begin{threeparttable}
\begin{tabular} {l c cccc c cccc c cccc c cccc}
\toprule
   &&    \multicolumn{4}{c}{MSR-VTT} &&  \multicolumn{4}{c}{MSVD} && \multicolumn{4}{c}{VATEX} &&  \multicolumn{4}{c}{YouCook2}\\
\cmidrule{3-6}\cmidrule{8-11}\cmidrule{13-16}\cmidrule{18-21}
Model    &&  B@4 & M & R & C &&  B@4 & M & R & C &&  B@4 & M & R & C &&  B@4 & M & R & C \\ 
\midrule
PMI-CAP\tnote{1}     && 42.1 & 28.7 & - & 49.4 && 54.6 & 36.4 & - & 95.1 &&  - & - & - & - && - & - & - & - \\
ORG-TRL\tnote{2}     && 43.6 & 28.8 & 62.1 & 50.9 && 54.3 & 36.4 & 73.9 & 95.2  &&  32.1 & 22.2 & 48.9 & 49.7 && - & - & - & - \\
OpenBook\tnote{3}    && 33.9 & 23.7 & 50.2 & 52.9 && - & - & - & - && 33.9 & 50.2 & 23.7 & 57.5 && - & - & - & - \\
CLIP4Caption\tnote{4} && 46.1 & 30.7 & 63.7 & 57.7 && - & - & - & - && - & - & - & - && - & - & - & - \\
SwinBERT\tnote{5} && 41.9 & 29.9 & 62.1 & 53.8 && 58.2 & 41.3 & 77.5 & 120.6 && 38.7 & 26.2 & 53.2 & 73.0 && 9.0 & 15.6 & 37.3 & 109.0 \\
LAVENDER\tnote{6}    && - & - & - & 60.1 && - & - & - & 150.7 && - & - & - & - && - & - & - & - \\
VIOLET-v2\tnote{7}   && - & - & - & 58.0 && - & - & - & 139.2 &&  - & - & - & - && - & - & - & - \\
HiTeA\tnote{8}   && - & - & - & 62.5 && - & - & - & 145.1 &&  - & - & - & - && - & - & - & - \\
GIT-large\tnote{9} && 48.7 & 30.9 & 64.9 & 64.1 && 75.8 & \bf 48.7 & \bf 85.5 & 162.9 && \bf 41.6 & 26.2 & \bf 54.3 & 72.5 && 7.5 & 14.4 & 34.9 & 98.3 \\
\midrule
\multicolumn{21}{l}{\emph{Larger models}} \\
GIT2 (5B)\tnote{9} && \underline{54.8} & \underline{33.1} & \underline{68.2} & \underline{75.9} && \underline{82.2} & \underline{52.3} & \underline{88.7} & \underline{185.4} && \underline{42.7} & \underline{28.8} & \underline{56.5} & \underline{94.5} && 9.4 & 15.6 & \underline{37.5} & \underline{131.2}     \\
Flamingo (80B)\tnote{10}   && - & - & - & - &&  - & - & - & - && - & - & - & 84.2 && - & - & - & 118.6 \\
\midrule
\ours{} (0.4B) && \bf 50.5 & \bf \underline{33.1} & \bf 66.8 & \bf 73.5 && \bf 75.9 & 47.7 & 85.0 & \bf 165.5 && 39.6 & \bf 27.2 & 54.2 & \bf 79.5 && \bf \underline{10.0} & \bf \underline{16.1} & \bf 37.4 & \bf 118.2 \\
\bottomrule
\end{tabular}

\begin{tablenotes}[para,raggedright]
\item[1] \citep{chen-etal-eccv2020-learning}
\item[2] \citep{org-trl}
\item[3] \citep{openbook}
\item[4] \citep{clip4caption}
\item[5] \citep{Lin_2022_CVPR_SwinBERT}
\item[6] \citep{lavender}
\item[7] \citep{violet-v2}
\item[8] \citep{hitea}
\item[9] \citep{wang2022git}
\item[10] \citep{alayrac2022flamingo}
\end{tablenotes}

\end{threeparttable}

\caption{Video Captioning results. Metrics are BLEU@4, METEOR, Rouge-L, and CIDEr, respectively.
The best results are in \textbf{bold}, while the best results including larger models are \underline{underlined}.}
\label{tab:main-results}
\end{table*}

We evaluate our \ours{} model on a range of video captioning benchmarks where the model needs to generate a text description for a given video clip.

\subsubsection{Datasets}
Following existing work~\cite{Lin_2022_CVPR_SwinBERT,wang2022git}, we evaluate on four widely used video captioning datasets: MSR-VTT, MSVD, VATEX, and YouCook2.

\noindent\textbf{MSR-VTT}~\cite{msrvtt} consists of 10K video clips, each annotated with 20 captions.
We adopt the standard split with 6.5K training, 500 validation, and 3K test videos.
For all datasets unless otherwise noted, our two-stage pre-trained model is fine-tuned on the training split, with the validation set used for model selection and ablation studies.
Final results on the test split are reported.

\noindent\textbf{MSVD}~\cite{chen-dolan-2011-collecting} is a dataset of 2K YouTube video clips, each annotated with around 40 human-written captions.
We use the standard train/validation/test split of 1.2K/100/670 videos.

\noindent\textbf{VATEX}~\cite{Wang_2019_ICCV_VATEX} contains 41K video clips and 20 ground-truth captions per video.
We report results on the public test set.

\noindent\textbf{YouCook2}~\cite{youcook2} is a collection of cooking videos covering 89 recipes.
It contains 15K video clips, each with one annotated caption.
We report results on the validation set following standard practice.
In our experiments, all models including ours only use the video content as input, and do not leverage additional information such as ASR-generated narrations.

\subsubsection{Results}
We compare \ours{} with a range of state-of-the-art video captioning models, and report results using four standard evaluation metrics: BLEU@4~\cite{papineni-etal-2002-bleu}, METEOR~\cite{banerjee-lavie-2005-meteor}, ROUGE-L~\cite{lin-och-2004-automatic}, and CIDEr~\cite{cider}.

As shown in Table~\ref{tab:main-results}, our \ours{} model substantially outperforms existing approaches (excluding much larger ones with multi-billion parameters), beating the best previous model by an average of \textbf{9.7 points} in CIDEr score.

\paragraph{Impact of Image-Text Pre-Training}
Table~\ref{tab:main-results} illustrates the importance of jointly pre-training the entire model on massive image-text data.
The two best performing models, \ours{} and GIT-large, both performed image-text pre-training over the entire encoder-decoder model, teaching it fundamental vision-language knowledge that serves as the foundation of video understanding.
This is in contrast with many earlier methods that either directly train on video data or only initialize the encoder with image-text pre-trained models.
The result that \ours{} and GIT-large beat the other methods, often by large margins, indicate the importance of image-text pre-training.

\paragraph{Impact of Intermediate Video-Text Pre-Training}
When we further compare the performance between \ours{} and GIT-large, which have roughly the same number of parameters, it reveals the benefit of adding an intermediate video-text pre-training stage.
While GIT directly fine-tunes the image backbone on downstream video captioning datasets, our model conducts an intermediate pre-training stage on large-scale video-text data, better adapting the image backbone to learning video-specific skills such as temporal reasoning.
As a result, \ours{} significantly outperforms GIT-large on all datasets under most metrics, achieving a new state of the art on video captioning.

In summary, the results in Table~\ref{tab:main-results} demonstrate the effectiveness of our two-stage pre-training framework, a new paradigm for training generative video-to-text models.

\subsection{Video Question Answering}\label{sec:exp:videoqa}
\begin{table}[ht]
\small
\centering
\begin{threeparttable}
\begin{tabular} {l rr}
\toprule
Model    & MSRVTT-QA    & MSVD-QA \\ 
\midrule
JustAsk\tnote{1}     & 41.5  & 46.3 \\
MV-GPT\tnote{2}      & 41.7  & - \\
MERLOT\tnote{3}      & 43.1  & - \\
VIOLET\tnote{4}      & 43.9  & - \\
All-in-One\tnote{5}  & 42.9  & 46.5 \\
GIT-large\tnote{6}   & 42.7  & 55.1 \\
\midrule
\multicolumn{3}{l}{\emph{Larger models}} \\
GIT2 (5B)\tnote{6}        & 45.6  & \underline{58.2} \\
Flamingo (80B)\tnote{7}   & \underline{47.4}  & - \\
\midrule
\ours{} (0.4B) & \bf 45.4 & \bf 55.5 \\
\bottomrule
\end{tabular}

\begin{tablenotes}[para,raggedright]
\item[1] \citep{justask}
\item[2] \citep{mv-gpt}
\item[3] \citep{zellers2021merlot}
\item[4] \citep{fu2021violet}
\item[5] \citep{wang2022allinone}
\item[6] \citep{wang2022git}
\item[6] \citep{alayrac2022flamingo}
\end{tablenotes}

\end{threeparttable}

\caption{VideoQA results. Numbers are (top-1) exact match accuracy.}
\label{tab:videoqa_results}
\end{table}

While our model is not specifically pre-trained on Video Question Answering (VideoQA) data, we fine-tune and evaluate on VideoQA to test the generalization of our approach.
In particular, we experiment on MSRVTT-QA and MSVD-QA~\cite{xu_etal_2017_videoqa}, two popular open-ended VideoQA datasets.
As many previous VideoQA models are encoder-only, one standard practice in the literature is to convert open-ended VideoQA to multi-choice VideoQA by predicting one of the answers in the training data, which is then solved by a classification model.
In contrast, our model is able to directly generate the open-ended text answer given the video and question, and we use beam search during inference similar to the case of video captioning.

The results are shown in Table~\ref{tab:videoqa_results}.
Our \ours{} model again outperforms all existing models (barring those with orders of magnitude more parameters), including the ones specifically designed and trained for VideoQA such as JuskAsk, which indicates that our model generalizes well across different video-to-text generation tasks.

\subsection{Implementation Details}\label{sec:exp:impl}

Our method is implemented using PyTorch~\cite{pytorch} and fairseq~\cite{ott-etal-2019-fairseq}, following the original OFA implementation.

For each video, the shorter side of the video frames are resized to 224 pixels, and we linearly sample 8 frames per video during both training and inference.
During intermediate video-text pre-training, for each training instance, one sample is generated for video captioning and one for video-to-text matching.
One sample is generated for frame order modeling for every eight instances.
For MSR-VTT and VATEX, SCST training~\cite{scst} is further performed during fine-tuning using the same learning rate following standard practice in the literature.

Detailed hyperparameters used in our experiments can be found in Table~\ref{tab:hyperparams} of Appendix~\ref{sec:appendix:impl}.

\section{Analysis and Discussion}\label{sec:analysis}
In this section, we present further discussions and insights of our model.
We first discuss how the optimal IPT strategy is selected with an empirical study (Section~\ref{sec:analysis:ipt_strategy}) and then discuss an ablation comparing the \ours{} and \ours{}-FiD architectures (Section~\ref{sec:analysis:arch}).
Finally, qualitative examples generated by our model can be found in Appendix~\ref{sec:appendix:examples}.

\subsection{Selecting the Optimal IPT Strategy}\label{sec:analysis:ipt_strategy}

\begin{table*}[ht]
\setlength{\tabcolsep}{0.4em}
\footnotesize
\centering
\begin{tabular} {l l l rrrr}
\toprule
& Video-Text IPT Data    & Video-Text IPT Tasks & MSR-VTT   & MSVD  & VATEX & YouCook2 \\
\midrule
1 & None    & None  & 66.4 & 173.7 & 78.1 & 99.7 \\
2 & S-MiT   & Captioning    & 69.3 & 189.2 & 81.3 & 99.2 \\
3 & S-MiT + Kinetics   & Captioning    & 69.8 & 183.7 & 82.3 & 97.1 \\
4 & S-MiT + Kinetics + HowTo100M   & Captioning    & 68.7 & 189.7 & 83.9 & 109.7 \\
5 & S-MiT + Kinetics + HowTo100M   & Captioning + Matching   & \bf 71.5 & 194.8 & 84.9 & 117.4 \\
6 & S-MiT + Kinetics + HowTo100M   & Captioning (S-MiT only) + Matching   & 69.3 & 190.4 & 82.7 & 105.4 \\
7 & S-MiT + Kinetics + HowTo100M   & Captioning + Matching + FOM (C+G) & 70.7 & 191.9 & 85.4 & 116.4 \\
8 & S-MiT + Kinetics + HowTo100M   & Captioning + Matching + FOM (C) & 70.8 & \bf 196.0 & \bf 85.7 & \bf 117.7 \\
\bottomrule
\end{tabular}

\caption{Ablation on Intermediate Video-Text Pre-Training strategies. Numbers are CIDEr scores on the validation set. (C) and (G) indicate contrastive and generative Frame-Order Modeling, as discussed in Section~\ref{sec:method:pft}. }
\label{tab:ipt_ablation}
\vspace{-3mm}
\end{table*}

As shown in Table~\ref{tab:ipt_ablation}, we conduct an extensive empirical study on various intermediate pre-training (IPT) strategies, focusing on different combinations of IPT data and tasks.

\paragraph{IPT data: More is (generally) better}
We first examine the impact of IPT data by training with the video captioning IPT task on different subsets of the three IPT datasets outlined in Section~\ref{sec:exp:ipt_data} (rows 1-4).
Comparing row 1 with the rest of the table, it becomes clear that our intermediate video-text pre-training substantially improves the model accuracy across all datasets.
Rows 2-4 indicates that adding more datasets in general helps with the model performance.
For instance, adding HowTo100M significantly improves the accuracy on YouCook2, likely due to their domain similarity.

\paragraph{IPT tasks: Captioning alone is not sufficient}
In Section~\ref{sec:method:pft}, we propose three IPT tasks: video captioning, video-text matching, and frame order modeling (FOM).
Comparing row 4 and 5, it shows that adding video-text matching consistently improves performance on all datasets compared to training with the captioning task alone.
This suggests that the contrastive pre-training objective is helpful (when combined with generative ones) even when the downstream task is generative (i.e.~video-to-text generation).
Pre-training with only the downstream task is not optimal as shown in our experiments.
In addition, row 8 indicates that adding the FOM task can further improve the accuracy (except for MSR-VTT), presumably because FOM teaches the model additional temporal reasoning skills that cannot be learned through image-text pre-training.
As a result, we adopt all three datasets and all three tasks (row 8) in our final \ours{} model.

\paragraph{Generative and Contrastive FOM}
We introduced FOM variants in Section~\ref{sec:method:pft}: generative and contrastive FOM.
In Table~\ref{tab:ipt_ablation}, row 7 uses both while row 8 only uses contrastive FOM.
We observe that adding the generative FOM results in lower performance overall, which we hypothesize is due to the fact that the model is trained to directly generate the frame order, an artifact not used in downstream tasks.
On the contrary, the model learn to encode temporal information through contrastive FOM without generating undesirable artifacts.

\paragraph{Captioning on S-MiT only}
On row 6, we further experimented with a variant where we conduct the video captioning task only on the S-MiT dataset.
Among the three IPT datasets, only S-MiT has high-quality human-written captions. Kinetics is a classification dataset with a limited set of pre-defined labels that are usually words or short phrases, while the text accompanying each video in the HowTo100M dataset is the automatically generated transcript of the narration in the video, which is not the same as the caption (video description) and may contain errors from the ASR system.
Therefore, row 6 explores a strategy where the video captioning task is only applied to the high-quality S-MiT dataset while the video-text matching is applied to all three datasets.
Interestingly, this strategy does not work as well as conducting both tasks on all datasets (row 5) and results in a performance degradation on all four downstream datasets.
Therefore, this strategy was not adopted in our final model.

\subsection{\ours{} vs.~\ours{}-FiD}\label{sec:analysis:arch}

\begin{table}[ht]
\setlength{\tabcolsep}{0.23em}
\footnotesize
\centering
\begin{tabular} {l l rrrr}
\toprule
IPT Arch    & FT Arch & MSRVTT   & MSVD  & VATEX & YC2 \\
\midrule
FiD    & FiD  & 69.6  & 185.6 & 84.0 & 107.4 \\
FiD    & \ours{}      & 68.3 & 185.4 & 83.8 & 109.2 \\
\ours{}        & \ours{}      & 70.8 & 196.0 & 85.7 & 117.7 \\
\bottomrule
\end{tabular}

\caption{\ours{} and \ours{}-FiD (FiD in the table) architectures. Validation CIDEr scores are shown.}
\label{tab:videoofa_vs_fid}
\end{table}

In Section~\ref{sec:method:arch}, we propose two architectures: our full \ours{} model, as well as a more memory-efficient \ours{}-FiD variant.
As shown in Table~\ref{tab:hyperparams} in Appendix~\ref{sec:appendix:impl}, the \ours{}-FiD variant can save up to 60\% GPU memory and thus enable the use of a batch size that is twice as large as the one used in \ours{} on the same hardware.

Table~\ref{tab:videoofa_vs_fid} compares their performance on video captioning using validation CIDEr scores.
We experiment with either variant in both IPT and fine-tuning, and observe that, unsurprisingly, \ours{} outperforms \ours{}-FiD across the board.
This indicates that it is beneficial to learn to reason over the entire video in the whole encoder-decoder model compared to only in the decoder.
This way, the encoder may also be able to learn video representations that are beyond static visual concepts, which appears to be more effective than restraining the encoder to learning local representations and perform joint reasoning only in the decoder.

We also explore using \ours{}-FiD during the more expensive IPT stage while using the more powerful \ours{} architecture during downstream fine-tuning, which yields similar results compared to using \ours{}-FiD in both stages.
This suggests that our IPT stage is critical for adapting the image backbone to learning video-specific skills, and allowing full self- and cross-attention in the IPT stage benefits more significantly to the model performance.
\section{Related Work}\label{sec:related_work}

\paragraph{Video-to-Text Generation}
Video Captioning is an important task in the field of CV and NLP, which aims to generate textual descriptions for videos that can summarize the content of the video~\citep{kojima-etal-2002-natural,li-etal-iccv2015-describing,venugopalan-etal-2015-translating}.
With the advent of deep neural networks, researchers largely rely on encoder-decoder~\citep[seq2seq,][]{seq2seq} networks for video captioning.
Early deep learning methods usually rely on offline-extracted video features~\cite{nayyer-etal-2019-spatio-temporal,pan-etal-2020-spatio-temporal,li2021value}, while more recent efforts propose end-to-end Transformer models that take raw video frames as input~\cite[][\emph{inter alia}]{clip4caption,Lin_2022_CVPR_SwinBERT}.

Video Question Answering (VideoQA) is another actively studied video-to-text generation task that involves answering questions about a given video~\cite{zeng2017leveraging,zhu2017uncovering,xu_etal_2017_videoqa}.
VideoQA is typically tackled by classifying the video-question input into a fixed vocabulary of answers~\cite{justask,zellers2021merlot,fu2021violet}, but we adopt a true open-ended setting where free-form answers are generated by the decoder.
Many of these video captioning and VideoQA models also leverage video-language pre-training to learn better representations as discussed in the next paragraph.

\paragraph{Video-Text Pre-Training}
Video-text pre-training is an actively studied research area where large-scale video-language data is leveraged to learn improved representations for various downstream tasks~\cite{videobert,li-etal-2020-hero,actbert}.
Some popular pre-training tasks include masked visual modeling~\cite{li-etal-2020-hero,fu2021violet}, masked language modeling~\cite{Su2020VL-BERT}, video-text matching~\cite{clipbert}, and frame order modeling~\cite{li-etal-2020-hero,zellers2021merlot}.
While most video-text pre-training works focus on encoder-only models for representation learning, some recent papers have proposed video-text pre-training for generative models that is suitable for tasks such as video-to-text generation~\cite{Luo2020UniVL,mv-gpt,lavender}.

\paragraph{Image-Text Pre-Training}
Similar to video-text pre-training, image-text pre-training aims to learn better vison-language representations from massive weakly supervised image-text data~\cite{tan-bansal-2019-lxmert,uniter}.
As more orders of magnitude more image-text data is available than video-text data, highly successful image-text models such as CLIP~\cite{pmlr-v139-radford21a} have been developed, which, via straightforward adaptations, can outperform video-text pre-trained models on video tasks~\cite{clip4clip,clip4caption}.
Several previous works also exist on learning generative image-text models through pre-training~\cite{vl-t5,unitab,wang-etal-2022-ofa,wang2022git}.
In this work, we adopt OFA~\cite{wang-etal-2022-ofa} as our backbone model, but our two-stage pre-training method is a general framework for generative video-text tasks that is compatible with improved generative image-text models as they become available in the future.

\section{Conclusion}\label{sec:conclusion}
In this paper we analyze a multitude of state-of-the-art video-to-text generation models and identify an important limitation that has eluded the researchers so far.
We propose a new two-stage pre-training framework to address this limitation that consists of an image-text pre-training and an intermediate video-text pre-training stage.
Through an extensive empirical study on the \emph{architectures} and \emph{training strategies} (Section~\ref{sec:analysis:ipt_strategy}) for the intermediate pre-training stage, we derive a new state-of-the-art video-to-text generation model, \ours{}, which significantly outperforms the best existing model by an average of 9.7 CIDEr points on four popular video captioning datasets.
\ours{} also achieves state-of-the-art performance on two open-ended Video Question Answering benchmarks despite not being pre-trained on VideoQA data, demonstrating the generalization of our model.

For future work, our two-stage pre-training framework can be extended beyond video-to-text generation by adding more IPT datasets and tasks, to build a more general multimodal generative video understanding model that can handle a variety of tasks such as video-to-text, text-to-video, video-to-image (highlight or poster creation), and video-to-video generation (trailer generation).
\section*{Acknowledgements}
We thank Satwik Kottur, Pedro Rodriguez, and Mahmoud Azab for helpful discussions throughout this project.

\bibliography{anthology,custom}

\clearpage

\onecolumn

\begin{appendices}
\section{Implementation Details}\label{sec:appendix:impl}

\begin{table*}[ht]
\centering
\begin{tabular} {l rrrr}
\toprule
Dataset & \#epochs & \#GPUs & batch size & learning rate \\
\midrule
IPT   & 10 & 32/16 & 4/8 & \num{1e-5} \\
\midrule
MSR-VTT & 15 & 8 & 10/24 & \num{1e-5} \\
MSVD & 15 & 4 & 10/24 & \num{5e-6} \\
VATEX & 15 & 16 & 10/24 & \num{1e-5} \\
YouCook2 & 20 & 8 & 10/24 & \num{1e-5} \\
MSRVTT-QA & 20 & 16 & 10/24 & \num{1e-5} \\
MSVD-QA & 20 & 8 & 10/24 & \num{1e-5} \\
\bottomrule
\end{tabular}

\caption{Hyperparameters used in our experiments. Per-GPU batch size is reported. For fields with a slash ($x/y$), $x$ is the valued used with the \ours{} architecture, while $y$ is for \ours{}-FiD.}
\label{tab:hyperparams}
\end{table*}

\section{Qualitative Examples of \ours{} Predictions}\label{sec:appendix:examples}

\begin{figure*}[h]
    \centering
    \includegraphics[width=\linewidth]{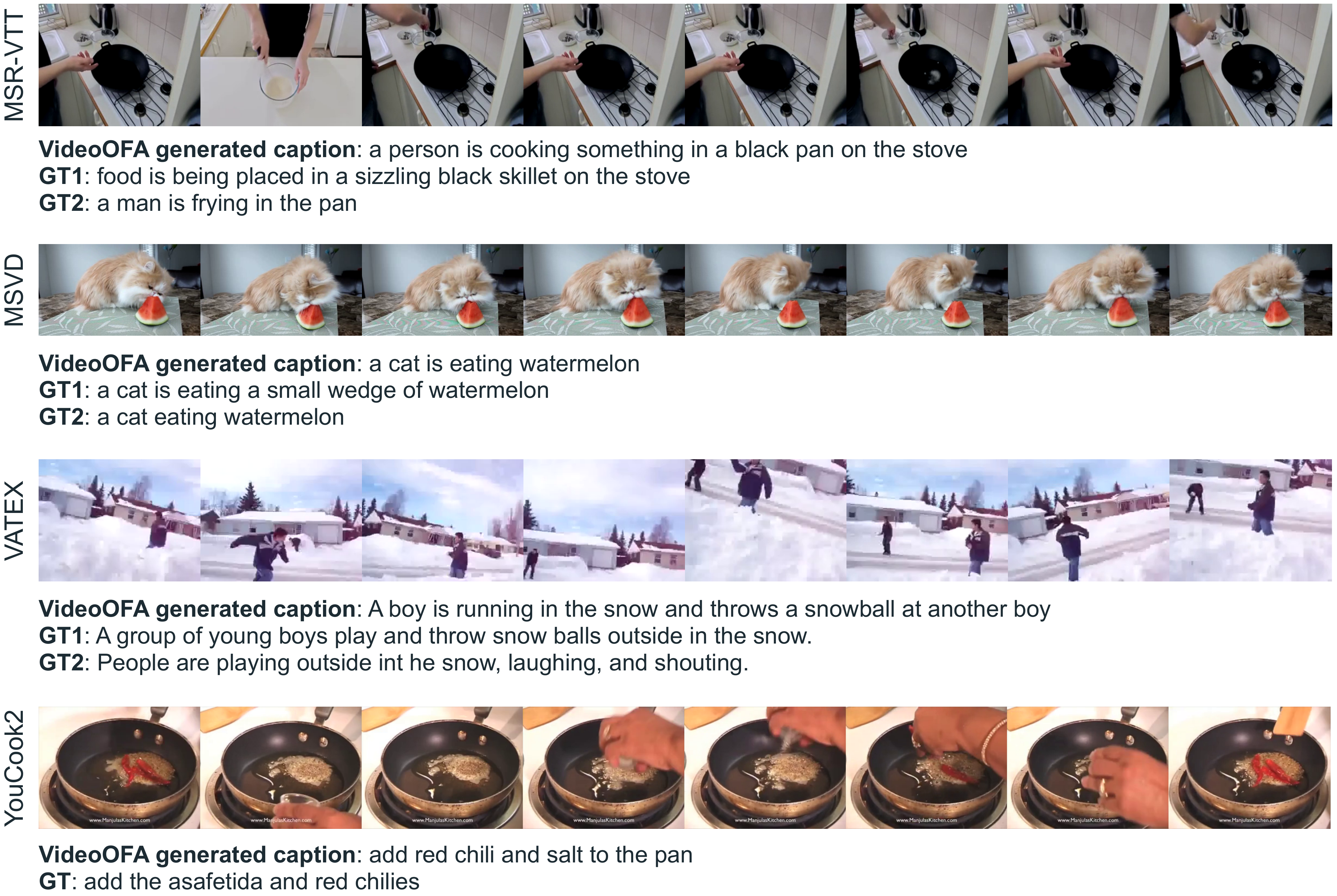}
    \caption{Qualitative examples generated by our VideoOFA model. Typos are from the original datasets.}
    \label{fig:examples}
\end{figure*}
\end{appendices}

\end{document}